\documentclass[10pt,twocolumn,letterpaper]{article}
\usepackage[final]{cvpr}  
\usepackage{caption}
\usepackage{placeins}
\usepackage{svg}
\usepackage[ruled,vlined,noend]{algorithm2e}
\SetKwComment{Comment}{/* }{ */}
\DontPrintSemicolon
\usepackage{adjustbox,booktabs,multirow} 
\usepackage{amsthm}
\usepackage{bbm}

\newtheorem{corollary}{Corollary}

\usepackage{booktabs}
\usepackage{multirow}

\usepackage[dvipsnames]{xcolor}

\definecolor{cvprblue}{rgb}{0.21,0.49,0.74}
\usepackage[pagebackref,breaklinks,colorlinks,allcolors=cvprblue]{hyperref}

\def\confName{CVPR}
\def\confYear{2026}

\title{DeepShare: Sharing ReLU Across Channels and Layers for Efficient Private Inference}

\author{
\begin{tabular}{cc}
Yonathan Bornfeld & Shai Avidan \\
\multicolumn{2}{c}{Tel Aviv University}
\end{tabular}
}

\begin{document}
\maketitle

\begin{abstract}

Private Inference (PI) uses cryptographic primitives to perform privacy preserving machine learning. In this setting, the owner of the network runs inference on the data of the client without learning anything about the data and without revealing any information about the model. It has been observed that a major computational bottleneck of PI is the calculation of the gate (i.e., ReLU), so a considerable amount of effort have been devoted to reducing the number of ReLUs in a given network. 

We focus on the DReLU, which is the non-linear step function of the ReLU and show that one DReLU can serve many ReLU operations. We suggest a new activation module where the DReLU operation is only performed on a subset of the channels (Prototype channels), while the rest of the channels (replicate channels) replicates the DReLU of each of their neurons from the corresponding neurons in one of the prototype channels. We then extend this idea to work across different layers.

We show that this formulation can drastically reduce the number of DReLU operations in resnet type network. Furthermore, our theoretical analysis shows that this new formulation can solve an extended version of the XOR problem, using just one non-linearity and two neurons, something that traditional formulations and some PI specific methods cannot achieve. We achieve new SOTA results on several classification setups, and achieve SOTA results on image segmentation.  \footnote{Our code will be made publicly available upon acceptance.}.

\end{abstract}    
\section{Introduction}
\label{sec:intro}

\begin{figure}[tbp]
    \centering
    \includegraphics[width=0.8\linewidth]{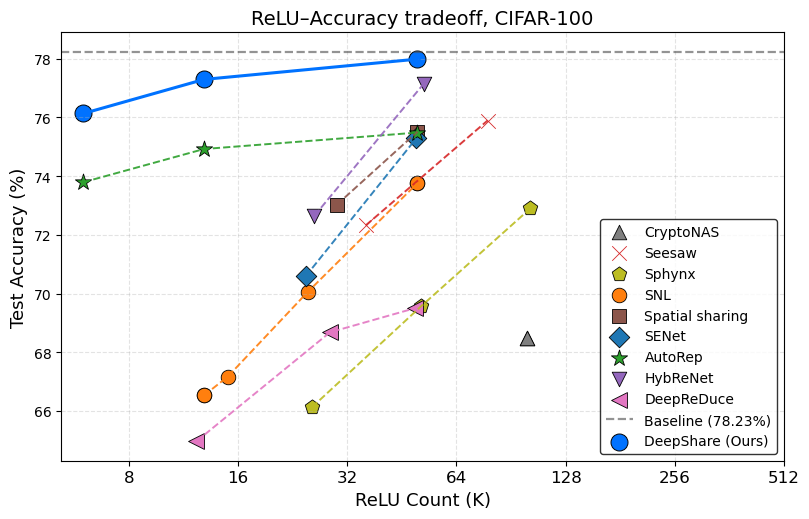}
    \\
    \caption{Our approach, DeepShare, achieves the Pareto frontier on
ReLU counts vs. test accuracy for CIFAR-100 using ResNet 18.}
    \label{fig:cifar100}
\end{figure}

With the rise of cloud based Machine Learning models (Machine-Learning-As-A-Service, MLaaS), there is a growing concern about the privacy of user data. Private Inference (PI) address this by using a protocol where the cloud provider (the model owner) and the user (the data owner) infer the result of the model on the data without leaking information to either side, other than letting the user know the final result.

PI can rely on homomorphic encryption to allow the model operate directly on encrypted data, but such schemes support only a limited set of operations. Alternatively, PI can be implemented using cryptographic protocols that require multiple rounds of communication.

A common observation in the literature is that in such protocols, the cost of linear operations (i.e., convolutions, fully connected layers) is fairly small, compared to the cost of the gate operation (i.e., ReLU). As a result, a growing number of researchers developed ways to reduce the number of gates in a given network. Our work falls into this category. Typically, reducing the number of gates involves finding a subset of the neurons that keep their non-linear gates (i.e., ReLU). The rest of the neurons are left without a gate (i.e., without a ReLU at all) or, in some cases, use a polynomial instead of ReLU. Instead of having a network where some of the neurons are linear (without a gate), we wish to share the non-linear gates with all the neurons of the network, in the hopes that it will increase its expressive power. This is also based on an analysis showing that ReLU gates of a neuron in a given spatial location are highly correlated across channels.

To share ReLU gates across neurons, we start by representing the ReLU operation as the product of its input and the DReLU of the input, where DReLU is a step function. This turns ReLU into a multiplication operator and pushes the gate decision to the DReLU. {\em DeepShare} shares DReLUs across channels within a layer and across layers within a block.
In DeepShare, each layer of the network is decomposed into prototype channels that behave similarly to channels in a regular network, and replicate channels. The activation of a replicate channel is the product of its input and the DReLU of neurons in a corresponding prototype channel. 

We conduct a number of experiments on different models and datasets, both for image classification and image segmentation. In all classification setups we outperform the current state of the art by a considerable margin, across many working points (for example, see figure \ref{fig:cifar100}), and in the segmentation setup we match the current state of the art. Then, we offer a theoretical explanation to the expressive power of our network representation. Specifically, we show that DeepShare can solve the classic XOR problem using just a single gate, that is shared across multiple neurons. \\
\indent To conclude, our main cotributions are:
\begin{itemize}
    \item We introduce a new approach for reducing PI inference cost through \emph{ReLU sharing} with our method, \textsc{DeepShare}.
    \item We establish new state-of-the-art results across multiple architectures and datasets.
    \item We provide a theoretical analysis that illustrates the expressiveness of \textsc{DeepShare}, suggesting how it may capture complex patterns despite operating under a substantially reduced ReLU budget.
\end{itemize}

\section{Related Works}
\label{sec:related}



The literature on Private Inference (PI) spans several complementary directions. One line of work focus on designing efficient inference protocols, for example through homomorphic encryption or by combining multiple cryptographic primitives. Another direction aims to develop models tailored to these protocols, seeking to maintain performance while enabling fast encrypted inference—either by proposing novel architectures or by adapting existing ones. Within this direction, a prominent sub-area, and the focus of our work, is the reduction of nonlinear operations—particularly ReLUs—which are computationally expensive in PI settings due to the substantial communication overhead they require.

\paragraph{PI Protocols:} These protocols focus on the fusion of cryptographic primitives with the basic building blocks of a neural network.

CryptoNets~\cite{gilad2016cryptonets} is one of the earliest efforts to address PI, by using homomorphic encryption.  SecureML~\cite{mohassel2017secureml} suggested a protocol for using three different cryptographic secret sharing protocols: Additive, Boolean, and Yao sharing. MiniONN~\cite{liu2017oblivious} and GAZELLE~\cite{juvekar2018gazelle} use additively homomorphic encryption to increase the communication efficiency of linear layers, while Blind faith \cite{khan2021blind} operates on homomorphic encrypted data using Chebyshev polynomials. FALCON~\cite{li2020falcon} improved  efficiency by running convolution in the frequency domain, and Chameleon~\cite{riazi2018chameleon} 
employs a Semi-honest Third Party dealer to generate Beaver's triplets in an offline phase. SecureNN~\cite{wagh2019securenn} suggested three-party computation (3PC) protocols for PI. This was later improved by CrypTFlow~\cite{kumar2020cryptflow} and Wagh {\em et al.}~\cite{wagh2020falcon}. HummingBird \cite{maeng2024accelerating} and Circa \cite{ghodsi2021circa} suggests methods to evaluate ReLU operations more efficiently, Crypten~\cite{knott2021crypten} complements these efforts by providing a software framework for private training and inference using secret sharing.

\paragraph{ReLU Pruning} 
ReLU operations are widely recognized as a major computational bottleneck in PI settings. Consequently, many methods aim to reduce the number of ReLUs while making the most effective use of those that remain. These approaches differ in how they decide which ReLUs to remove and in the alternative operators they use in place of the removed nonlinearities. \\
Some works \cite{jha2023deepreshape} point out that FLOPS from other operations can also contribute significantly to the inference time. Nevertheless, ReLU pruning remains important, as it can be combined with these complementary techniques to further reduce overall computational cost.\\ 
A family of methods leverage Network Architecture Search (NAS) algorithms to determine the networks structure and nonlinearities. within those methods, DELPHI \cite{srinivasan2019delphi} replaces the removed ReLU with a quadratic function, CryptoNAS \cite{ghodsi2020cryptonas} replace the ReLU neurons with linear neurons, and it and Sphynx \cite{cho2022selective} modify other operations inside the block to apply ReLU on a smaller number of neurons. Other methods, such as SENet \cite{kundu2023senet} and DeepReDuce \cite{jha2021deepreduce}, use importance measurements in order to decide which ReLUs to prune. SENet \cite{kundu2023senet} uses weight-pruning approaches to describe the importance of ReLU operations in different layers, and iteratively updates a binary mask indicating which ReLU to keep, while DeepReDuce \cite{jha2021deepreduce} uses three different stages of reducing ReLUs using different strategies, requiring several manual design choices.

A prominent work in the field is selective Network Linearization (SNL) \cite{cho2022selective}, using a differential optimization process for choosing which neurons to linearize. they do so by assigning a learnable parameter per neuron, governing whether it uses ReLU or not, and optimizing these parameters along with finetuning the model. AutoRep \cite{Peng_2023_ICCV} extends this work by introducing a 2-nd order polynomial nonliearity for neurons that do not have a ReLU operation assigned to them, and use STE for the optimization of the parameters governing the ReLU assingment. Gorski {\em et al.} \cite{gorski2023securingneuralnetworksknapsack} is closest to our method, suggesting ReLU sharing in the {\em spatial} domain, utilizing the observation that spatially close neurons display high correlation in their ReLU gate values. their solution however requires a discrete optimization step prior to network training.

\section{Method}
We begin by analyzing ReLU gate correlations, motivating the idea of sharing them. then we present two methods for sharing ReLU gates across neurons: \textbf{Channel Sharing}, which share them across channels within the same layer, and \textbf{Layer Sharing}, which share them across different layers. Then, we outline the training process and introduce the concept of a \textbf{transitional GELU phase}.

\paragraph{Notations:} We define ReLU as the product of the input and a gate, termed DReLU (Derivative of ReLU): 
\begin{equation}
\mathrm{DReLU}(x) =
\begin{cases}
1, & \text{if } x \ge 0, \\
0, & \text{otherwise.}
\end{cases}
\end{equation}
\begin{equation}
\mathrm{ReLU}(x) = x \cdot \mathrm{DReLU}(x)
\end{equation}
\subsection{ReLU gate analysis}
We investigate whether the ReLU gates of a trained network exhibit cross-channel correlations at corresponding spatial locations. If such correlations exist, they might indicate the potential for cross-channel ReLU-gate sharing.

To assess this, we record the ReLU gate values of a trained ResNet-18 on CIFAR-100 across \(N\) examples (we use \(N = 4096\)). For each spatial position, this yields \(N\) activation vectors of dimension \(d\), where \(d\) is the number of channels in the layer. To quantify cross-channel dependencies, we perform PCA on these vectors and compute their effective dimension (ED), following \cite{del2021effective}:
\[
\mathrm{ED} = \frac{\left( \sum_{i=1}^{d} \lambda_i \right)^{2}}{\sum_{i=1}^{d} \lambda_i^{2}},
\]
where \(\lambda_1, \ldots, \lambda_d\) are the PCA eigenvalues.

To isolate the role of spatial structure, we repeat the same procedure after randomly shuffling the spatial positions independently within each channel. This preserves each channel’s marginal distribution while removing spatial alignment across channels. The resulting ED values serve as a baseline capturing correlations that are \emph{not} tied to shared spatial positions.

\begin{figure}
    \centering
    \includegraphics[width=0.98\linewidth]{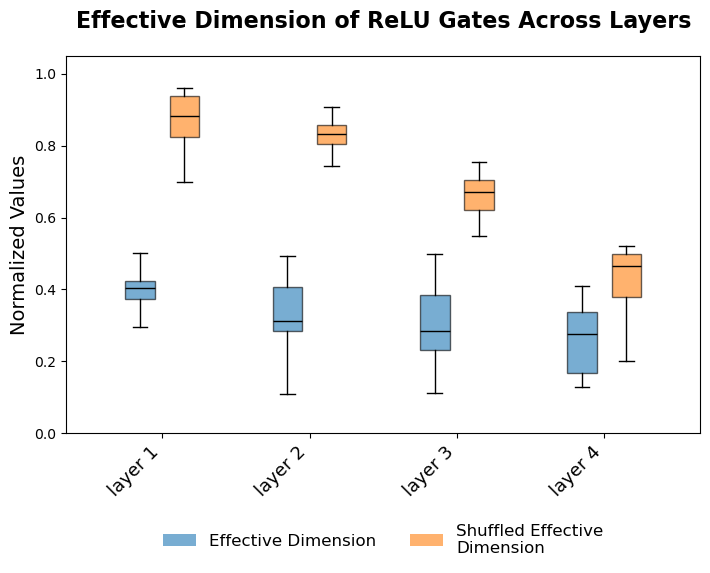}
    \caption{\textbf{Effictive Dimension of ReLU gates across channels:} Boxplots of the normalized effective dimension across four layers. For each layer, the blue boxes show the effective dimension at corresponding spatial positions, The orange boxes show the effective dimension obtained after spatial shuffling within each channel. both normalized by the actual dimension, meaning the values are between 0 and 1, with smaller numbers indicating that the gates are correlated and have an effective dimension much smaller than the real one. 
}
    \label{fig:dim}
\end{figure}

We observe, in Figure~\ref{fig:dim} that the effective dimension at corresponding spatial positions is approximately \(0.3\text{--}0.4\) of the full dimensionality, and is substantially lower than in the shuffled baseline. This indicates meaningful cross-channel correlations that arise specifically at shared spatial locations, supporting the motivation behind sharing ReLU gates across channels.
\subsection{Channel Sharing Module}
\begin{figure}
    \centering
    \includegraphics[width=0.98\linewidth]{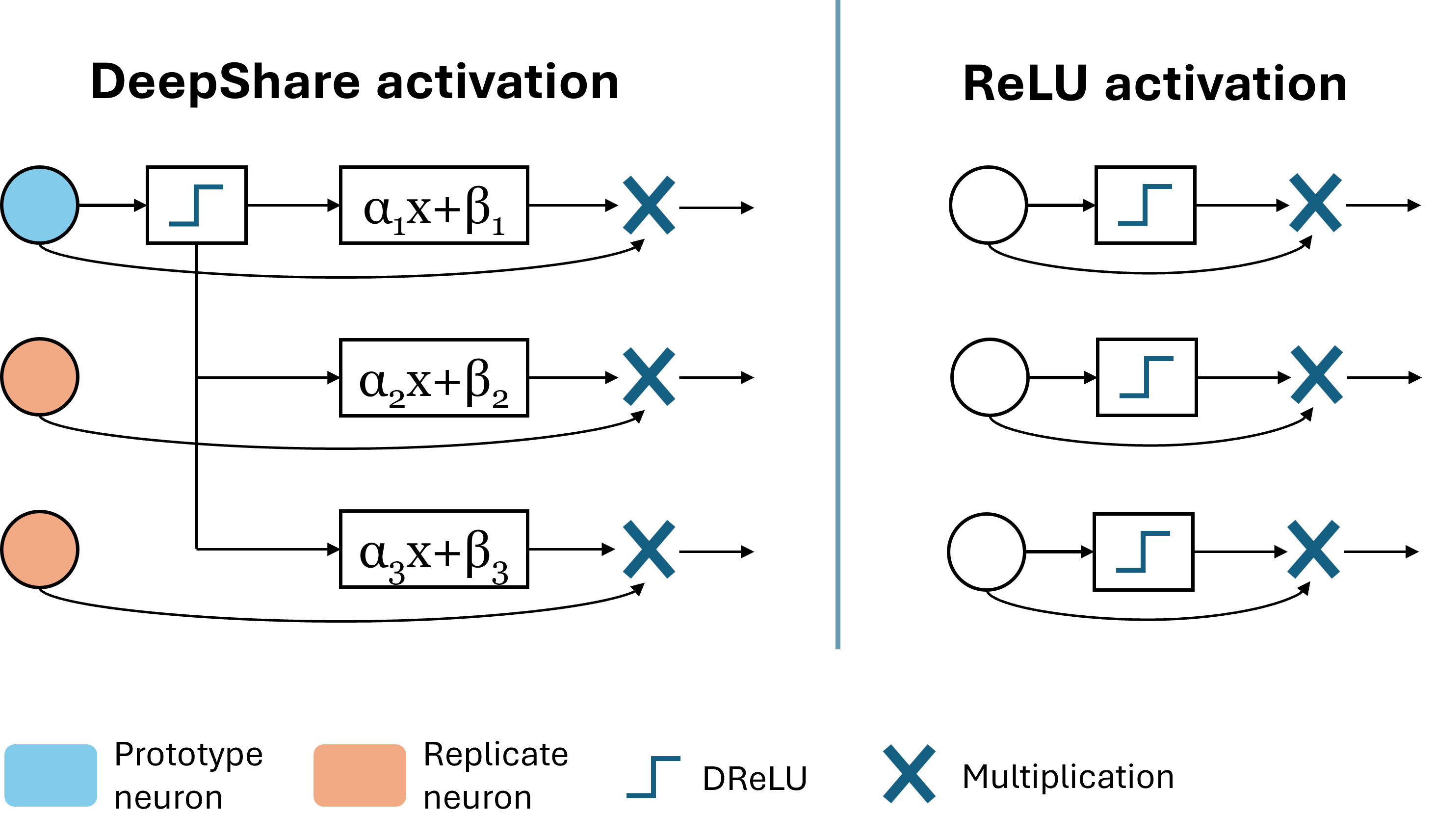}
    \caption{{\bf DReLU Sharing:}
    DeepShare shares DReLU across channels of the same layer. (Right) in standard ResNet activation, the ReLU is the product of the input and a DReLU (gate) operation on the input. (Left) In our DReLU sharing scheme, channels in a layer are partitioned into {\em prototype} and {\em replicate} channels. The ReLU of a prototype neuron is the product of its input and the DReLU of the input (top row, blue neuron). In contrast, the ReLU of a replicate neurons (bottom two rows) is the product of their input (orange) and the DReLU of the corresponding product neuron (blue). All three neurons share the DReLU of the prototype neuron, thus reducing overall number of gates in the network. The 1D affine transformation adds flexibility and expressive power to the network.}
    \label{fig:method}
\end{figure}

We focus here on networks that consists of layers, where each layer consists of multiple channels (i.e., ResNet or CNN type networks).
Given layer $l$ with $C$ channels, we split it into $P$ prototype and $R$ replicate channels, such that $C=P+R$. By convention, the prototype channels are the first channels in the layer. The gate operation is applied on the neurons of the prototype channels and shared with the neurons of the replicate channels. The key concept is given in Figure~\ref{fig:method}.
\paragraph{Affine Transformation} 
Simply replicating the gate from prototype to replicate neuron is susceptible to some failure cases. For example, in the extreme case that the number of prototype channels $P=1$, then if its gate is close (i.e., its DReLU is zero) then all the replicate neurons that depend on it will be zeroed out as well. Using an affine transformation can avoid this scenario, since the values will not necessarily zero out. Combining ReLU sharing and adding affine transformation we arrive at the following definition:
\begin{equation}
    \mathrm{sharedReLU}(x^{c}) = 
    x^{c} \cdot 
    \left(\alpha^{c} \cdot \mathrm{DReLU}(x^{\pi(c)}) + \beta^{c} \right)
\end{equation}
where $x^{c}$ is a neuron in channel $c$ and $\pi({c})$ is a mapping from the replicate to prototype channels in a naive balanced way. In case the mapping $\pi$ is the identity, we are back to the standard definition of a ReLU. The scalars $\alpha^{c}$, $\beta^{c}$ are the weight and bias that are learned per layer and channel. Similarly, in case $\alpha^{c}=1$ and $\beta^{c}=0$ we are back to the original definition of ReLU. We found that adding this transformation introduces greater diversity among the replicate neurons. We also note that it is a relatively inexpensive operation, requiring only a scalar multiplication and addition.

\paragraph{DReLU Budget}
Since we only use DReLU from prototype channels, our total DReLU count for the model is $\sum_{l=1}^{L}P^lh^lw^l$, instead of $\sum_{l=1}^{L}C^lh^lw^l$ in the original network, where $P^l, C^l$ are the number of prototype channels and the total number of channels of layer $l$, respectively. Similarly, $w^l, h^l$ denote the width and hight of a layer $l$.
In order to determine the amount of prototype channels and replicate channels at each layer, we need to create a DReLU layer budget. for simplicity, we used the basic SNL \cite{cho2022selective} method to extract the ReLU amount per each layer, and use it to determine $P^l$ and $Q^l$. 

\paragraph{Layer Sharing Module}
We extend the idea of ReLU sharing to operate across layers, whenever a sequence of consecutive layers shares identical spatial and channel dimensions. Specifically, we partition such sequences into groups and apply a procedure analogous to channel-wise sharing. In each group, the channels of the first layer are divided into prototype and replicate channels. For every subsequent layer in the group, each channel is aligned with its corresponding channel in the first layer, copies its DReLU activation, and then applies a learnable affine transformation (unique to each channel and layer). The layer budget of DeepShare is determined by SNL. Specifically, we run SNL and set the DeepShare budget per layer as a function of the number of ReLUs found by SNL for that layer. We then scale the budget of all layers to reach the required total budget.

\subsection{Training Process}
\begin{figure}
\includegraphics[width=0.98\linewidth]{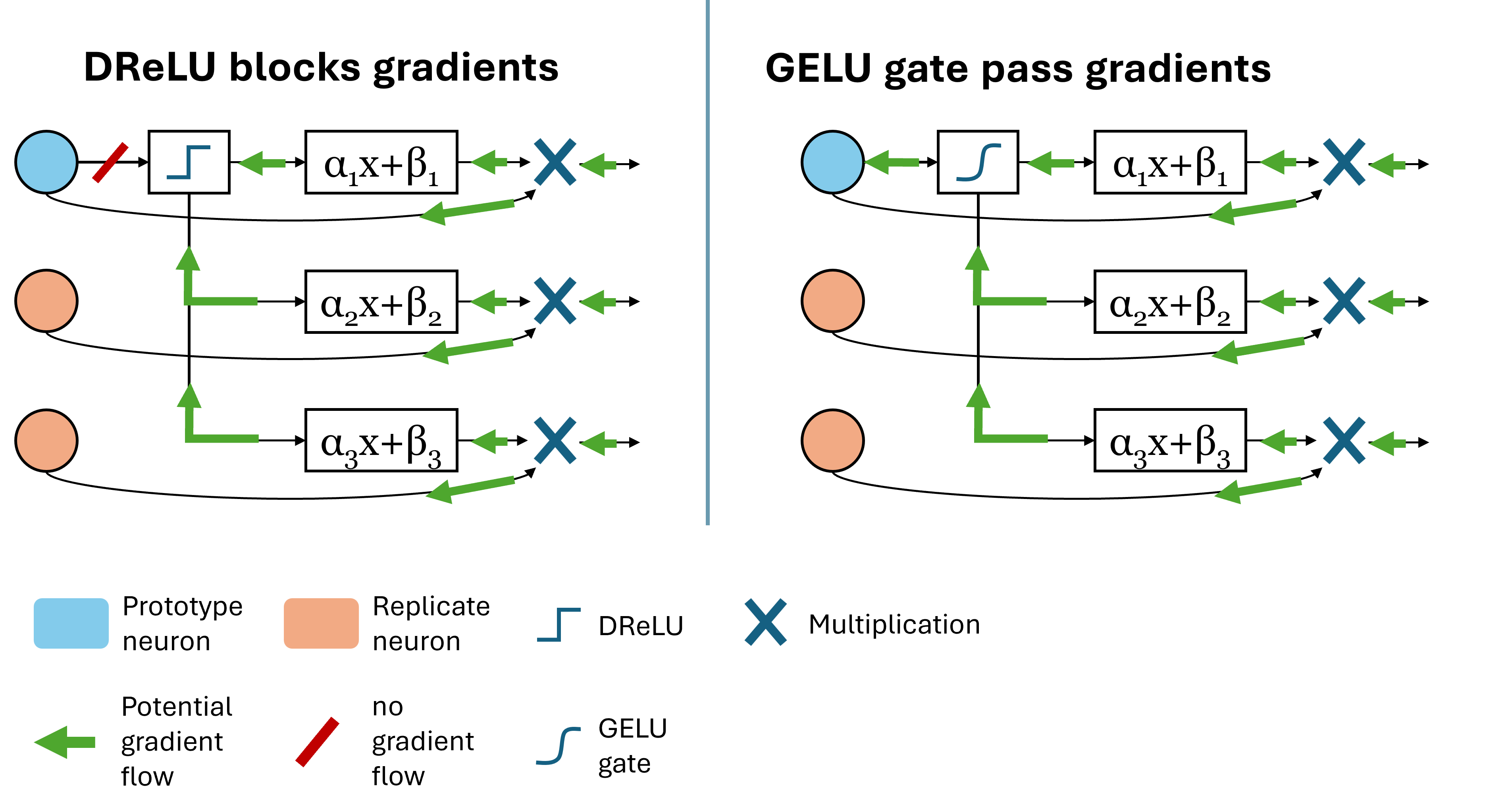}
    \caption{{\bf Gradient Flow:} An illustration of the gradient flow between the different operations during back propagation.  (left) DReLU has a derivative of zero so it blocks (the red cut on the edge between the prototype neuron and its DReLU) the flow of the gradients from the replicate neurons to the prototype neuron. (Right) gating component of GELU has a non-zero derivative so the gradients are not blocked. This is why it is advantageous to first train with it before switching to DReLU.}
    \label{fig:grad}
\end{figure}

A key challenge in the training process is that the DReLU function has a derivative of zero (or it is not differentiable at $0$). because of that, even though the DReLU value of prototype neurons affect multiple replicate neurons, the prototype neuron will not get gradients related to this effect through back propagation (as illustrated in \ref{fig:grad}), which might lead to a sub-optimal optimization process. In order to overcome this challenge, we start the training with a \textbf{transitional GELU phase}. a
GELU (Gaussian Error Linear Unit) is defined as: 
$$\mathrm{GELU}(x) = x P(X \le x) =x\Phi(x)$$
where $P$ is the standard normal distribution, and $\Phi(x)$ is its cumulative distribution function. We note that $\Phi$ in GELU acts as a gate modifier by multiplication. We can use at $\Phi$ as a proxy of DReLU, that can supply gradients for the learning process. Practically, we start by training with $\Phi(\gamma \cdot x)$ instead of DReLU, for some hyperparameter scalar $\gamma$ that controls how sharp the function curve is. By doing so, we allow back propagation of the affect of the prototype neurons on the replicate neurons, back to the prototype neurons. Afterwards, we gradually replace the GELU gate with DReLU (by replacing them channel by channel for all of the layers simultaneously), while using knowledge distillation from the GELU version of the model. An overview of the algorithm is given in Alg~\ref{alg:drelu_progressive}.

\begin{algorithm}[t]
\caption{DeepShare training process}
\label{alg:drelu_progressive}
\KwIn{Training data $\mathcal{D}$, total ReLU budget $B$}
\KwOut{Trained model parameters $\theta^*$}

\Comment{Step 1: DReLU budget}
Create per-layer DReLU budgets using SNL allocation\;
Determine prototype and replicate channel counts accordingly\;

\Comment{Step 2: Pretraining phase}
Train a DeepShare model with GELU activations for the prototypes channels\;

\Comment{Step 3: Progressive ReLU substitution}
\While{not all prototype channels use ReLU}{
    \ForEach{layer $l$}{
        Switch a subset of prototype channels from GELU to ReLU\;
    }
    Train the model for one epoch on $\mathcal{D}$\;
}

\Comment{Step 4: Final finetuning}
Finetune the resulting\;
\Return $\theta^*$\;
\end{algorithm}

\section{Experiments}
\paragraph{Cryptographic setup}
For our experiments, following \cite{Peng_2023_ICCV}, we adopt a two-party secure computation protocol for MLaaS. We assume a semi-honest adversarial model \cite{choi2019hybrid, zhang2013picco}, in which parties adhere to the protocol but may attempt to infer private information from observed communications. Our implementation is based on the standard secret-sharing scheme used in CrypTen \cite{knott2021crypten}, which can, in principle, support two-party computation through the use of off-line oblivious transfer.

We evaluate our method on multiple models and datasets, including classification and segmentation tasks. \\
for classification, we experiment with ResNet18 on CIFAR100 and Tiny ImageNet, and with Wide-Resnet 22-8 on CIFAR100. for segmentation, we experiment with DeepLab3 with MobilenetV2 backbone on ADE20K.

We compare against a number of methods. Some focus on the efficient use of cryptographic blocks in the protocol, like CryptoNAS~\cite{ghodsi2020cryptonas}, Seesaw \cite{li2024seesaw}, and Sphynx~\cite{cho2022sphynx}. Others focus on reducing the number of ReLUs, as we do. This includes SNL~\cite{cho2022selective}, AutoReP~\cite{Peng_2023_ICCV}, SENet~\cite{kundu2023senet}, HybReNet \cite{jha2023deepreshape}, and spatial sharing of Gorski \etal~\cite{gorski2023securingneuralnetworksknapsack}. In the following experiments we report the accuracy of the different methods for different values of ReLU counts. In addition, we report a normalized score of the ratio of accuracy to the number of ReLUs, as custom in other works ~\cite{Peng_2023_ICCV, cho2022selective}.

\begin{table}[t]
\centering
\caption{CIFAR-100 Comparison, ResNet 18 based methods and novel architectures}
\setlength{\tabcolsep}{4pt}            
\renewcommand{\arraystretch}{1.1}      
\begin{adjustbox}{max width=\columnwidth} 
\begin{tabular}{p{1.4cm} lcccc}        
\toprule
 & \textbf{Methods} & \textbf{\#ReLUs (K)} & \textbf{Acc. (\%)} &  \textbf{Acc./\#ReLU} &\\
\midrule\midrule
\multirow{3}{*}
& Baseline       & 491.52 & 78.23 & 0.16\\
\midrule\midrule
\multirow{9}{*}{\rotatebox[origin=c]{90}{$\mathbf{50K \le ReLU \le 100K}$}}
& CryptoNAS      & 100   &  68.5  & 0.68 \\
& Seesaw          & 78 & 75.89 & 0.97 \\
& Sphynx         & 51.2  & 69.57  & 1.36 \\
& SNL            & 49.9  & 73.75  & 1.48   \\
& Spatial sharing & 50  & 75.51  & 1.51   \\
& SENet & 49.6  & 75.28  & 1.51   \\
& AutoRep        & 50    & 75.48  & 1.51 \\
& HybReNet       & 52    & 77.14 & 1.48 \\
& DeepShare (Ours)  & 49.9 & \textbf{77.98}   & \textbf{1.56} \\
\midrule\midrule
\multirow{7}{*}{\rotatebox[origin=c]{90}{$\mathbf{10K < ReLU < 50}$}}
& Seesaw          & 36 & 72.32 & 2.00 \\
& Spatial sharing & 30  & 73.03  & 2.43   \\
& HybReNet       & 26    & 72.65 & 2.79 \\
& SENet           & 24.6  & 70.59  & 2.87   \\
& SNL            & 12.9 & 66.53 & 5.15 \\
& AutoRep        & 12.9 & 74.92 &  5.80  \\
& DeepShare (Ours)  & 12.9 & \textbf{77.29} &  \textbf{5.99}     \\
\midrule\midrule
\multirow{2}{*}{\rotatebox[origin=c]{90}{$\mathbf{< 10K}$}}
& AutoRep        & 6 & 73.79    &  12.29  \\
& DeepShare (Ours)  & 6 & \textbf{76.13}    &  \textbf{12.69} \\
& & & & \\

\bottomrule
\end{tabular}
\end{adjustbox}
\label{tab:cifar100ResNet18}
\end{table}

Table~\ref{tab:cifar100ResNet18} shows the results of running ResNet-18 based methods and novel architectures on the cifar-100 dataset. The baseline ResNet-18 achieves accuracy of $78.23\%$ (when trained with knowledge distillation) with about $500k$ ReLUs, whereas we achieve accuracy of $76.13\%$ with $6K$ DReLUs, reducing the number of non-linearities by two orders of magnitude with a drop of about $2\%$ in accuracy. 
We outperform all other methods across all working points (49.9K ReLUs, 12.9K ReLUs, 6K ReLUs), often by a large margin. For example, we outperform AutoRep \cite{Peng_2023_ICCV} by more than $2\%$ across all working points, despite the fact that AutoReP replaces the linear neurons with polynomial ones for better accuracy. Furthermore, our lowest ReLU setup, 6K, achieves higher accuracy than their highest ReLU setup of 50K. reaching $0.65\%$ accuracy improvement with a factor of 8.3X less ReLUs.

\begin{table}[t]
\centering
\caption{CIFAR-100 Comparison, with Wide-ResNet 22-8}
\setlength{\tabcolsep}{4pt}            
\renewcommand{\arraystretch}{1.1}      
\begin{adjustbox}{max width=\columnwidth} 
\begin{tabular}{p{1.4cm} lcccc}        
\toprule
 & \textbf{Methods} & \textbf{\#ReLUs (K)} & \textbf{Acc. (\%)} & \textbf{Acc./\#ReLU}  & \\
\midrule
\multirow{3}{*}
& Baseline       & 1359.87 & 80.2 & 0.06\\
\midrule\midrule
\multirow{5}{*}{\rotatebox[origin=c]{90}{\scriptsize $\mathbf{150k\le ReLU}$}}
& SENet          & 240 & 79.3 & 0.33 \\
& SENet          & 180 & 79.02 & 0.44 \\
& SNL            & 180 & 77.65 & 0.43 \\
& AutoRep        & 150 & 78.38 & 0.52 \\
& DeepShare (Ours)           & 150 & \textbf{79.43} & \textbf{0.53} \\
\midrule\midrule
& SNL            & 120 & 76.35 & 0.63 \\
& AutoRep        & 120 & 77.56 & 0.64 \\
& DeepShare (Ours)  & 120 & \textbf{79.37} & \textbf{0.66} \\
\midrule\midrule
& DeepShare (Ours)  & 60 & \textbf{78.87} & \textbf{1.31} \\
\bottomrule
\end{tabular}
\end{adjustbox}
\label{tab:cifar100WideResNet22}
\end{table}

We observe similar behavior when running Wide-ResNET-22-8 on Cifar-100. See Table~\ref{tab:cifar100WideResNet22}. 
We achieve an accuracy of $78.87\%$ using 60K DReLUs, compared to an accuracy of $80.2\%$ of the base model that uses nearly 1.4M ReLUs.
Again, we outperform all other methods across all working points. 

These results beg the question of how to measure the power of a network? For example, we have that Wide-ResNet 22-8 with roughly 1.4M parameters and only 60K gates achieves an accuracy of $78.87\%$. This accuracy is {\em higher} than that of a standard ResNet 18 architecture that consists of roughly 500K parameters and 500K gates, and achieves an accuracy of $78.23\%$. We leave this question for future research.

\begin{table}[t]
\centering
\caption{Tiny-Imagenet Comparison, ResNet 18 based methods and novel architectures}
\setlength{\tabcolsep}{4pt}            
\renewcommand{\arraystretch}{1.1}      
\begin{adjustbox}{max width=\columnwidth} 
\begin{tabular}{p{1.4cm} lcccc}        
\toprule
 & \textbf{Methods} & \textbf{\#ReLUs (K)} & \textbf{Acc. (\%)} &  \textbf{Acc./\#ReLUs} &\\
\midrule\midrule
\multirow{3}{*}
& Baseline       & 1966.08 & 65.48 & 0.03\\
\midrule\midrule
\multirow{9}{*}{\rotatebox[origin=c]{90}{$\mathbf{190K \le ReLU}$}}
& HybReNet          & 327     & 64.92 & 0.20\\
& SENet          & 298     & 64.96  & 0.22 \\
& Sphynx         & 204.8   & 53.51 & 0.26 \\
& DeepReduce     & 196.6   & 57.51 & 0.29 \\
& SNL            & 198.1   & 63.39 & 0.32 \\
& SENet          & 142     & 58.9 & 0.41 \\
& HybReNet       & 104 & 58.9 & 0.56 \\
& AutoReP        & 190     & 64.32 & 0.34 \\
& DeepShare (Ours) & 190     & \textbf{65.2} & \textbf{0.35} \\
\midrule\midrule
\multirow{4}{*}{\rotatebox[origin=c]{90}{\scriptsize $\mathbf{50K ... 60K}$}}
& HybReNet       & 52 & 54.46 & 1.05 \\
& SNL            & 59.1 & 54.24 & 0.92 \\
& AutoRep        & 55 & 63.69 & 1.16 \\
& DeepShare (Ours)  & 55 & \textbf{64.4} & \textbf{1.17} \\
\midrule\midrule
& AutoRep        & 30 & 62.77 & 2.09 \\
& DeepShare (Ours)  & 30 & \textbf{62.8} & 2.09 \\

\bottomrule
\end{tabular}
\end{adjustbox}
\label{tab:TinyResNet18}
\end{table}

Table~\ref{tab:TinyResNet18} reports the results of running ResNet-18 on the Tiny imagenet dataset (where images are $64 \times 64$ pixels). Our method achieves an accuracy of $62.8\%$, compared to $65.48\%$ of the original network, while requiring just $30K$ DReLUs, compared to nearly 2M ReLUs in the original network, a drop of nearly two orders of magnitude in the number of gates required. As before, we observe that our method outperforms all other methods across all working points (See the supplementary material for the full comparison figure).
\begin{table}[t]
\centering
\caption{ADE20K Comparison, with MobileNetV2 with DeepLabV3.}
\setlength{\tabcolsep}{4pt}            
\renewcommand{\arraystretch}{1.1}      
\begin{adjustbox}{max width=\columnwidth} 
\begin{tabular}{lccc}
\toprule
\textbf{Methods} & \textbf{\#ReLUs (K)} & \textbf{mIoU (\%)} \\
\midrule
Baseline         & 85262 & 34.08 \\
Spatial sharing  & 2558  & \textbf{33.23} \\
DeepShare (Ours) & 2556  & 33.22 \\
\bottomrule
\end{tabular}
\end{adjustbox}
\label{tab:ADE20k}
\end{table}

In the last main experiment we apply DeepShare to the problem of image segmentation. Specifically, we use MobileNetV2 with DeepLabV3 to segment the ADE20K images. The typical size of these images is $512 \times 512$. See Table~\ref{tab:ADE20k}. The only other method reported in the literature is that of Gorski \etal \cite{gorski2023securingneuralnetworksknapsack}. We perform comparably.

\subsection{Ablations}
We evaluated the contribution of each of the components of our method and report results in Table~\ref{tab:ablations}. This ablation was conducted on CIFAR-100 with ResNet-18 and a budget of 6K DReLUs. Specifically, we ablated the following steps:

\begin{itemize}

\item {\bf No Affine transformation} 
We use our modules without the per-channel affine transformation, which means only sharing the DReLU.

\item {\bf No Layer sharing} we use only channel sharing, without layer sharing.

\item {\bf No Budget}
We use a fixed ratio of prototype channels for each layer, instead of relying on SNL to determine the number of prototype channels per layer.

\item {\bf No GELU phase} 
We run the training process without the transitional GELU phase. meaning, training with ReLU from the beginning.

\item {\bf No Replicates} We use a standard network, where
we keep the prototype channels but remove the replicate channels. This maintains the same amount of non-linearities.


\end{itemize}

As can be seen, each component contributes to the overall performance of the method. Of particular interest is the last row of the table that shows the accuracy in case we remove all replicate channels. This results in a trimmed ResNet with a considerably smaller number of channels per layer (but the same number of non-linearities as our method). However, there is a sharp drop in performance from $76.13\%$ of our full method to only $56.08\%$, which indicates the importance of including the replicate layers. We investigate possible reasons for this gap in Section~\ref{sec:Theory}.

\begin{table}[t]
\centering
\caption{Ablations, on CIFAR-100, Resnet 18, with 6K ReLU.}
\setlength{\tabcolsep}{4pt}            
\renewcommand{\arraystretch}{1.1}      
\begin{adjustbox}{max width=\columnwidth} 
\begin{tabular}{l c}
\toprule
\textbf{Methods} & \textbf{Acc. (\%)} \\
\midrule
Full method              & 76.13 \\
No affine transformation & 73.51 \\
No layer sharing         & 73.39 \\
No Budget                & 73.31 \\
No GELU phase            & 69.44 \\
No Replicates            & 56.08 \\
\bottomrule
\end{tabular}
\end{adjustbox}
\label{tab:ablations}
\end{table}

We have conducted a latency test on our algorithm, Using the CrypTen framework \cite{knott2021crypten}, running on two computers with Nvidia RTX A5000, connected to the same network with a bandwidth of 1GB. since latency measurements can vary significantly depending on factors like computational power, network availability, etc, we compare ourselves to AutoRep~\cite{Peng_2023_ICCV} who also used the CrypTen framework.
we compare the latency speedup factor, measured as the ratio between the latency of the model and the latency of the baseline version of that model, with all of the ReLU operations. both measured on the same computers (we use AutoRep reported measurements).
As can be seen, our result are slightly worst than AutoRep for all cases, which can be explained also by the setup gaps mentioned above.

\begin{table}[t]
\centering
\caption{{\bf Latency:} We compare the latency speedup factor of DeepShare Vs. AutoRep on various datasets and using different networks (compared to the standard model, higher is better).}
\setlength{\tabcolsep}{4pt}            
\renewcommand{\arraystretch}{1.1}      
\begin{adjustbox}{max width=\columnwidth} 



\begin{tabular}{l c c c}
\toprule
\textbf{Setup} & \textbf{\#ReLUs (K)} 
               & \multicolumn{2}{c}{\textbf{Latency Speedup}} \\
               &                    & \textbf{AutoRep} & \textbf{DeepShare} \\
\midrule

\multirow{3}{*}{CIFAR-100, ResNet 18}
 & 50    & 4.92 & 4.78 \\
 & 12.9  & 7.30 & 6.76 \\
 & 6     & 8.01 & 7.64 \\
\midrule\midrule

\multirow{2}{*}{CIFAR-100, Wide-ResNet 22-8}
 & 150   & 5.26 & 4.10 \\
 & 120   & 5.87 & 4.44 \\
\midrule\midrule

\multirow{2}{*}{Tiny-ImageNet, ResNet 18}
 & 55    & 8.63 & 6.81 \\
 & 30    & 9.60 & 8.25 \\
\bottomrule
\end{tabular}
\end{adjustbox}
\label{tab:latency}
\end{table}



\section{Theoretical Analysis}
\label{sec:Theory}
We aim to demonstrate that DeepShare is inherently more expressive than standard or SNL~\cite{cho2022selective} like networks with the same amount of ReLU gates, thereby providing preliminary explanation of its abilit to capture complex real-world patterns under a limited ReLU-gate budget. In standard networks, each neuron has its own gate (i.e., ReLU). SNL-type networks have mix of ReLU gates and linear neurons without gates. In DeepShare, one gate can be used by multiple neurons. In what follows, we will focus on a version of the XOR problem and show that only DeepShare can solve it using a {\em single} gate, while SNL cannot.

\begin{figure}[tbp]
    \centering
    \includegraphics[width=0.8\linewidth]{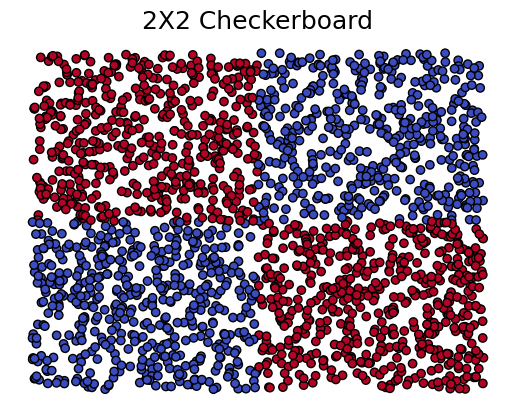}
    \\
    \caption{{\bf The XOR problem:} We wish to solve this classic XOR problem using just a {\em single} gate. We show that DeepShare can solve it using a single gate.}
    \label{fig:decision_boundary}
\end{figure}

\paragraph{Problem Definition.}
We consider the “$2\times2$ checkerboard” ~\cite{konyushkova2017learning, sun2014analysis}, an extension of the classic XOR task. See Figure~\ref{fig:decision_boundary}. The input space is the two-dimensional plane, and each point is labeled according to
\[
h(x_1, x_2)
    = \mathbbm{1}\!\left[\operatorname{sign}(x_1) \neq \operatorname{sign}(x_2)\right],
\]
where $\mathbbm{1}[\cdot]$ denotes the indicator function.
For a given model family, we ask whether it can represent this function using only one DReLU operation. This problem is intentionally chosen for its nonlinearity, which makes it challenging to solve using a single DReLU operation. We will show that our model succeeds within these limitations, while conventional and SNL models do not.

\paragraph{Constructive Proof.}
Our goal is to construct a network with a single DReLU operation that exactly reproduces $h(x_1,x_2)$.  
We use a single hidden layer with one \emph{prototype} neuron and one \emph{replicate} neuron. The key idea is that the function $$g(x_1,x_2)=-x_1x_2$$ solves this task, since $$\operatorname{sign}(x_1) \neq \operatorname{sign}(x_2) \iff -x_1x_2>0.$$
Utilizing the special definition of our replicate neuron activation, we will construct with our model a similar function $$f(x_1,x_2)=-\operatorname{sign}(x_1)x_2$$ that also meets the same criteria.

\begin{figure}[htbp]
    \centering
    \includegraphics[width=0.8\linewidth]{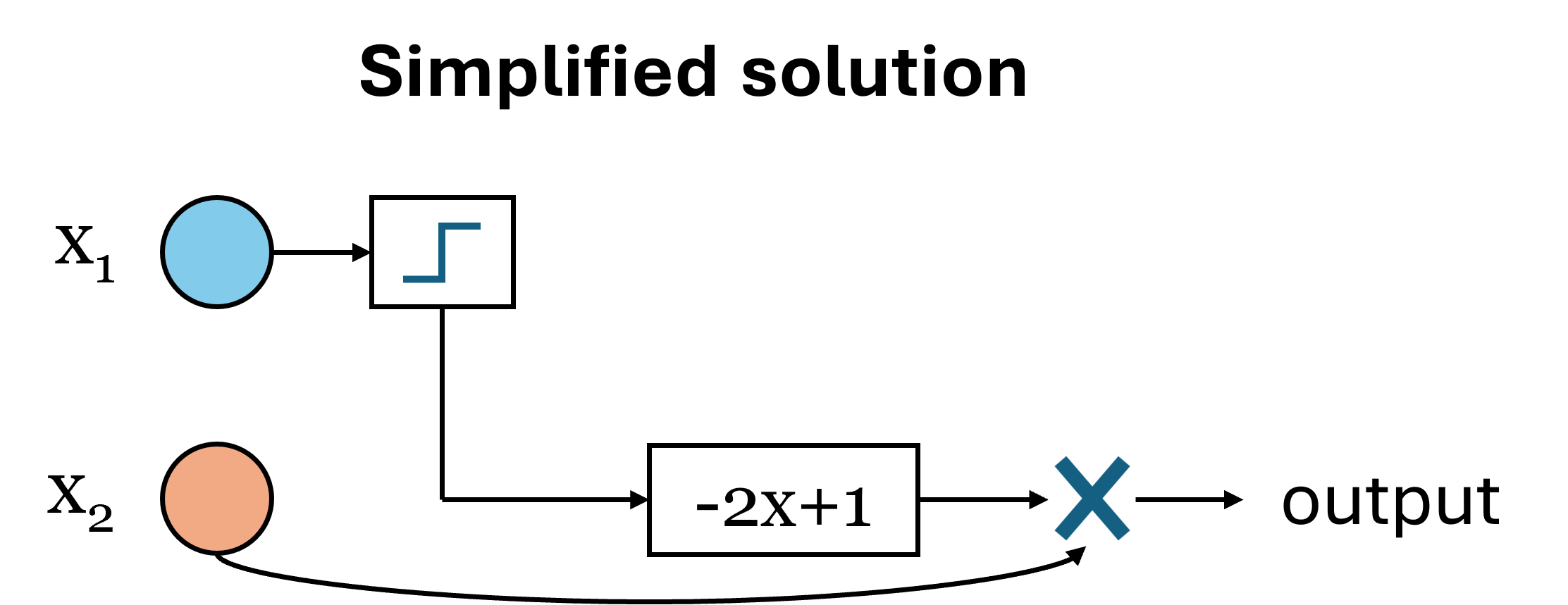}
    \\
    \caption{{\bf ReLU sharing solution to the XOR problem:} We show a solution to the XOR problem that requires just a single non-linearity. $(x_1,x_2)$ are the coordinates of the point to be classified. The DReLU of $x_1$ (the prototype, blue neuron) is passed through an affine function and multiplied with $x_2$, the replicate neuron (orange). See Equation~\ref{eq:xor_drelu}.} 
    \label{fig:visual_proof}
\end{figure}

Consider Figure~\ref{fig:visual_proof}, and observe that the activation of $x_2$ is the product of $x_2$ and the DReLU of $x_1$ (after an affine transformation). This lets us partition the 2D plane vertically (using DReLU($x_1$)) and horizontally (using $x_2$). \\
Formally, since we don't need the first standard linear layer to do any operations, we set it's weights and biases to be:
\[
W^1 =
\begin{bmatrix}
1 & 0 \\
0 & 1
\end{bmatrix},
b^1 =
\begin{bmatrix}
0 \\
0
\end{bmatrix}
\]

In order to convert $DReLU(x_1)$ to $-\operatorname{sign}(x_1)$, we choose the affine parameters $\alpha_2 = -2$ and $\beta_2 = 1$. \\
finally, our output will be the activation value of $x_2$, so we can set the second linear layer weights and biases to be 
\[W^2 =
\begin{bmatrix}
0 \\
1
\end{bmatrix},
b^2 = 0
\]
The model function can be described as:
\begin{equation}
f(x_1, x_2) = x_2 \cdot (-2\,\operatorname{DReLU}(x_1) + 1)=-\operatorname{sign}(x_1)x_2
\label{eq:xor_drelu}
\end{equation}
Since $\operatorname{sign}(x_1) \neq \operatorname{sign}(x_2) \iff f(x_1,x_2)>0$, the network exactly reproduces the desired checkerboard pattern using only a single DReLU operation.

This solution highlights the fact that sharing the DReLU values allows the network to separate the nonlinearity from the neuron value, and increase expressiveness.

\begin{figure}[htbp]
    \centering
    \includegraphics[width=0.8\linewidth]{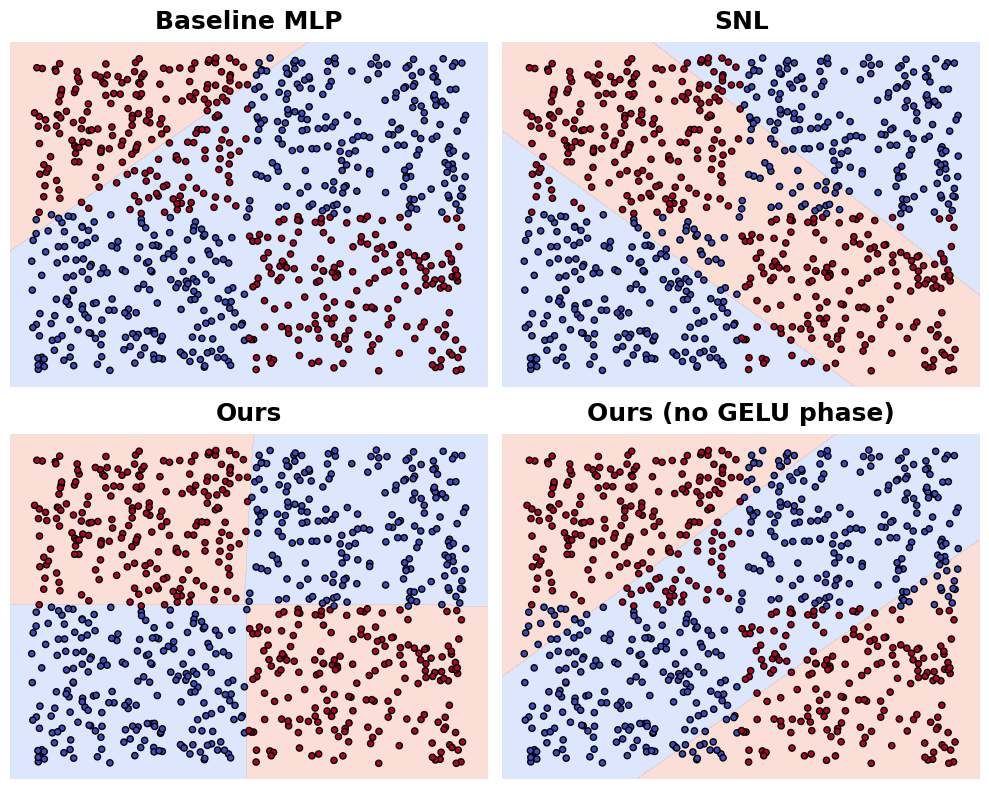} \\
    \caption{{\bf Training on the XOR problem:} We compare four different training methods for the XOR problem. (top-left) A single ReLU network, (top-right) SNL~\cite{cho2022selective} style network that combines one ReLU neuron with multiple linear neurons, (bottom right) Running our ReLU sharing training algorithm without the GELU phase, all fail to converge. (bottom left) Running our ReLU sharing algorithm with a transitional GELU phase converged to the correct solution.}
    \label{fig:xor_results}
\end{figure}

In addition, as seen in Figure~\ref{fig:xor_results}, we empirically reach an approximation of the required solution. This shows that the expressiveness benefit is not only theoretical but can be achieved by optimization. In addition, as further ablation for the transitional GELU phase, we note that training DeepShare with ReLU from the start does not converge to the correct solution, highlighting the importance of propagating the gradients through the sharing operation so the model can utilize the DReLU operations better.

\paragraph{SNL and MLP fail to express the task}
A conventional MLP with a single hidden neuron and a monotonic activation can only induce a single linear decision boundary, insufficient to represent the alternating structure of the checkerboard function.
Similarly, for SNL, we show in the supplementary the follwing:
 \begin{corollary}\label{cor:snl_limitation}
An SNL model with a single hidden layer, a single ReLU neuron, and any number of linear neurons, will have either no decision boundary, or one of the following decision boundaries: a single line, two parallel lines, or a piecewise linear with two pieces
\end{corollary}
Using Corollary~\ref{cor:snl_limitation}, we see that none of the possible decision boundaries of the model can solve the task. 

\section{Conclusions}
\label{sec:conclusions}
We propose a novel direction for increasing models efficiency for Private Inference (PI) that is based on ReLU sharing across channels and layers. In our formulation, termed {\em DeepShare}, the ReLU is represented as the product of its input and a its gate operation, termed DReLU. The DReLU of a neuron in a channel can then be shared with corresponding neurons in other channels of the same layer in the network. Towards this end, we group the channels in each layer into prototype channels, where DReLUs are computed, and replicate channels, that rely on the DReLUs of the prototype channels. This idea was further extended to work across layers.

We evaluated our approach on a number of network architectures and a number of datasets and found that it consistently outperform the current state of the art, sometimes by a considerable margin. For example, in the case of running ResNet 18 on CIFAR-100 with 12.9K ReLUs, we improved the accuracy of SOTA from $74.92\%$ to $77.29\%$, a jump of more than $2\%$. This is just $1\%$ below the accuracy of the base model that requires 491K ReLUs.
We then suggest a theoretical explanation to the performance of our network by showing that DeepShare can solve a version of the XOR problem using a {\em single} non-linearity, which standard ReLU networks and SNL networks cannot do. 
\clearpage
{
\small
\bibliographystyle{ieeenat_fullname}
\bibliography{main}
}

\clearpage
\setcounter{page}{1}
\maketitlesupplementary
\appendix
\paragraph{Overview.}
This supplementary material offers extended theoretical proofs, implementation clarifications, and further analyses supporting the main paper.
In \cref{sec:snl_proof}, we present a proof of the expressiveness limitations of single-layer SNL models, complementing the theoretical claims in the main text.  
\Cref{sec:run_details} details the full training setup used in our experiments, including optimization hyperparameters, data augmentation, and the gradual transition from GELU to ReLU activations.  
In \cref{sec:layer_sharing}, we expand on the layer–sharing mechanism and describe how prototype–replicate mappings are extended across groups of layers.  
\Cref{sec:additional_figs} provides additional figure of empirical results of the Pareto frontier for Tiny-ImageNet.  
Finally, \cref{sec:affine_analysis} offers an analysis of the learned affine gate parameters, providing initial observations on how prototype and replicate channels utilize the shared gating structure.


\section{Theoretical Proof}
\label{sec:snl_proof}
We provide proof for corollary 1 in the main paper. 

\begin{figure*}[tbp]
    \centering
    \includegraphics[width=0.8\linewidth]{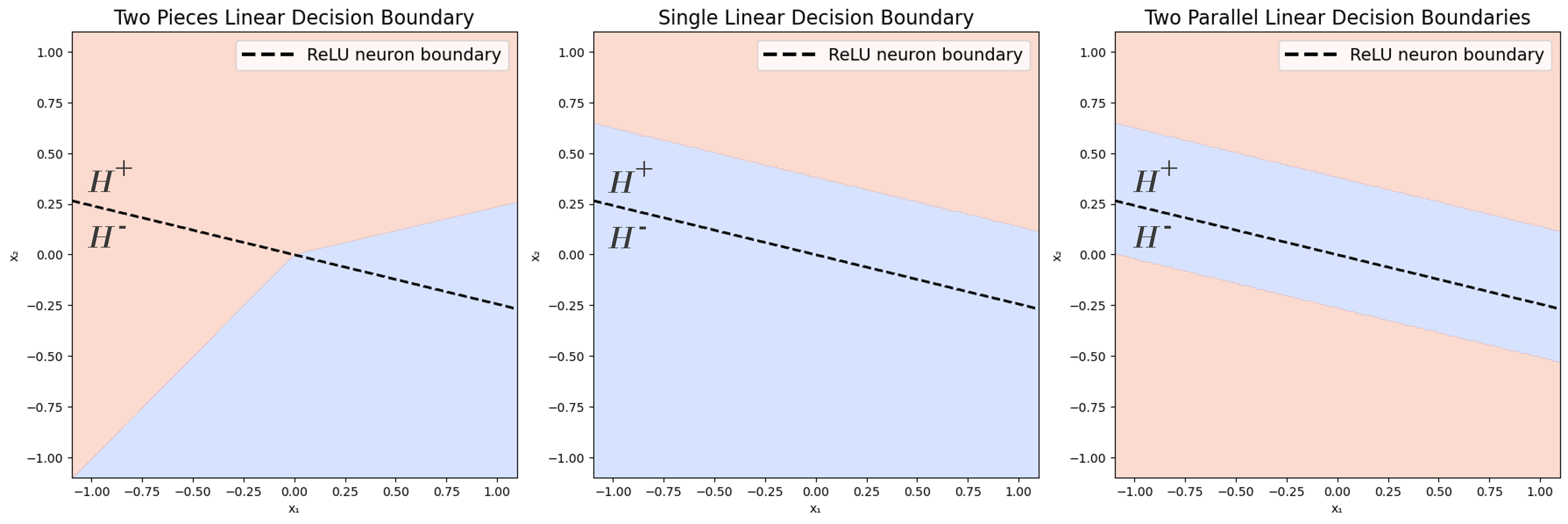}
    \caption{
\textbf{Possible decision boundaries induced by SNL models.} 
The dashed line denotes the activation boundary of the ReLU neuron, referred to in the proof as $H$, with $H^{+}$ and $H^{-}$ being the two half-planes it partitions. 
Within each half-plane, the function is affine and can generate at most one decision line. 
The three scenarios shown are: 
\emph{left}: a piecewise-linear boundary consisting of two segments that meet on $H$; 
\emph{middle}: a single decision line; 
\emph{right}: two parallel decision lines.
}
\label{fig:snl_boundary}
\end{figure*}

\begin{em}

An SNL model with a single hidden layer, a single ReLU neuron, and any number
of linear neurons will have either no decision boundary, or one of the
following decision boundaries: a single line, two parallel lines, or a
piecewise linear curve with two linear pieces. See Figure~\ref{fig:snl_boundary} for visualization.
\end {em}

\begin{proof}
We consider a binary classifier of the form
\[
\hat{y}(x) = \operatorname{sign}(f(x)), \qquad x \in \mathbb{R}^2,
\]
where $f$ is implemented by a single-hidden-layer SNL network with one ReLU
hidden unit and any number of linear hidden units.

\paragraph{Step 1: Functional form of the SNL model.}
Let the input be $x \in \mathbb{R}^2$. Since there's only a single hidden layer, All the linear neurons are equivalent to a single linear neuron in terms of expressivity, so we can write the SNL score function as
\begin{equation}\label{eq:snl_form}
f(x) = a\,\sigma(w^\top x + b) + v^\top x + c,
\end{equation}
with parameters $w,v \in \mathbb{R}^2$, and $a,b,c\in\mathbb{R}$, and the RELU function $\sigma(z)=max\{0, z\}$.

\paragraph{Step 2: Closed half-planes and affine structure.}
Define the ReLU hyperplane
\[
H := \{x : w^\top x + b = 0\}
\]
and the two closed half-planes
\[
H^+ := \{x : w^\top x + b \ge 0\}, \qquad
H^- := \{x : w^\top x + b \le 0\}.
\]
Note that $H^+ \cup H^- = \mathbb{R}^2$ and $H^+ \cap H^- = H$.

On each of these closed half planes, the ReLU takes a fixed linear form, so $f$ becomes affine:
\begin{itemize}
    \item On $H^-$, $\sigma(w^\top x + b)=0$, hence
  \[
  f(x) = v^\top x + c, \qquad x\in H^-.
  \]
\item On $H^+$, $\sigma(w^\top x + b)=w^\top x + b$, hence
\[
\begin{aligned}
f(x)
&= a(w^\top x + b) + v^\top x + c \\
&= (v + aw)^\top x + (c + ab),
\qquad x \in H^+.
\end{aligned}
\]
\end{itemize}

\paragraph{Step 3: Shape of the zero set.}
The decision boundary is the zero level set
\[
Z := \{x \in \mathbb{R}^2 : f(x) = 0\}.
\]

From the affine expressions above, define the two lines
\[
\begin{aligned}
L_- &:= \{x : v^\top x + c = 0\}, \\
L_+ &:= \{x : (v + a w)^\top x \\
    &\qquad\quad + \,(c + ab) = 0\}.
\end{aligned}
\]
Then
\[
Z \cap H^- = L_- \cap H^-, \qquad
Z \cap H^+ = L_+ \cap H^+.
\]
Hence
\[
Z = (L_- \cap H^-) \cup (L_+ \cap H^+).
\]
Also, in the the overlapping hyperplane the two functions agree, and therefore we get
\[
H\cap L_- = H \cap L_+
\]

We now enumerate all possible configurations of the lines $L_-$ and $L_+$ (illustrated  in Figure \ref{fig:snl_boundary}):
\begin{enumerate}
    \item Both intersections $L_- \cap H^-$ and $L_+ \cap H^+$ are empty.  
    In this case $f$ has no zeros, and therefore no decision boundary appears.

    \item Exactly one of the intersections is empty.  
    Then all zeros lie on the remaining line, yielding a single linear decision boundary.

    \item One of the lines is parallel to $Z:=H^- \cap H^+ = \{x : w^\top x + b = 0\}$.  
    The other line cannot be non-parallel, since a non-parallel line would intersect $Z$ in a single point, whereas the parallel one would not; this contradicts the fact that the two affine expressions of $f$ coincide on $Z$.  
    Hence in this situation the decision boundary is either a single line or a pair of parallel lines.

    \item Both $L_- \cap H^-$ and $L_+ \cap H^+$ are nonempty and the two lines are not parallel.  
    Each line then meets $Z$ at a unique point, and because the affine pieces agree on $Z$, these intersection points must coincide.  
    Consequently, the decision boundary consists of two linear pieces meeting at a single point, i.e., a two-piece piecewise-linear boundary.
\end{enumerate}

These cases exhaust all possibilities and match exactly the forms stated in the corollary.
\end{proof}

\section{Run Details}
\label{sec:run_details}
\paragraph{Hyperparameters}

Table~\ref{tab:hyperparams} summarizes the hyperparameters used across all experiments.  
We employ the SGD optimizer with momentum and weight decay, along with learning-rate schedules similar to prior work~\cite{Peng_2023_ICCV, cho2022selective, gorski2023securingneuralnetworksknapsack}.
For data augmentation, we applied random cropping, horizontal flipping, and color jitter.

For the scaling factor of the GELU phase, $\alpha \in \mathbb{R}$ in the gated form $\Phi(\alpha x)$, we set $\alpha = 1$ for all classification tasks.  
For the segmentation experiments with MobileNetV2, we observed that the smaller input magnitudes caused the activation to behave overly linearly.  
To mitigate this effect, we used a sharper gate by setting $\alpha = 4$.  

Regarding knowledge distillation, as discussed earlier, the GELU-phase model served as the teacher for the subsequent two phases, using a temperature of $4$ for all classification tasks.  
During the GELU phase itself, we employed a standard model with GELU activations as the teacher.  
Knowledge distillation was not applied in the segmentation experiments. \\
For the XOR experiments, we sampled 800 points uniformly from $[-1,1]^2$ and trained each model for 5{,}000 epochs using SGD (learning rate 0.1, batch size 32).

\paragraph{Switch from GELU to ReLU}
After completing the GELU phase, we gradually convert the network’s gates from GELU to ReLU.  
The conversion occurs progressively over several epochs, with more channels switching to ReLU as training advances.

Within each layer, channels are converted in a fixed order: we begin from the last channel and move backward. Thus, prototype channels are converted last.

During the transition, replicates switch independently and are not influenced by whether their prototypes have already changed from GELU to ReLU.  
Both GELU and ReLU gates are precomputed and replicated across all channels, and the process simply selects, for each channel, which version to use.  
When a channel is scheduled to switch, its GELU gate is replaced with its corresponding ReLU gate, while the rest of the computation remains unchanged.

\begin{table*}[t!]
\centering

\caption{Hyperparameter settings used across models, datasets, and training phases. When a scheduler is used, the LR column shows the full LR range ($\text{start} \rightarrow \text{end}$). when a linear schedular was used, it end epoch was $75\%$ of the total epochs of the phase}
\label{tab:hyperparams}
\begin{adjustbox}{max width=\textwidth}
\begin{tabular}{l l l c c c c c c}
\toprule
\textbf{Model} & \textbf{Dataset} & \textbf{Phase} 
& \textbf{Epochs / Steps} & \textbf{Batch} 
& \textbf{LR (start→end)} & \textbf{Scheduler} 
& \textbf{Momentum} & \textbf{Weight Decay} \\ 
\midrule

ResNet18 & CIFAR-100 & GELU phase
& 100 epochs & 128 & $0.05 \rightarrow 5\times10^{-4}$ 
& Linear & 0.9 & 0.001 \\

ResNet18 & CIFAR-100 & Switch to ReLU 
& 150 epochs & 128 & $1\times10^{-3}$ 
& None & 0.9 & 0.001 \\

ResNet18 & CIFAR-100 & Finetune 
& 120 epochs & 128 & $1\times10^{-3} \rightarrow 1\times10^{-5}$ 
& Linear & 0.9 & 0.001 \\
\midrule

WRN-22-8 & CIFAR-100 & GELU phase 
& 100 epochs & 128 & $0.05 \rightarrow 5\times10^{-4}$ 
& Linear & 0.9 & 0.001 \\

WRN-22-8 & CIFAR-100 & Switch to ReLU 
& 150 epochs & 128 & $1\times10^{-3}$ 
& None & 0.9 & 0.001 \\

WRN-22-8 & CIFAR-100 & Finetune 
& 120 epochs & 128 & $1\times10^{-3} \rightarrow 1\times10^{-5}$ 
& Linear & 0.9 & 0.001 \\
\midrule

ResNet18 & Tiny-ImageNet & GELU phase
& 100 epochs & 128 & $0.05 \rightarrow 1\times10^{-5}$ 
& Cosine & 0.9 & 0.001 \\

ResNet18 & Tiny-ImageNet & Switch to ReLU
& 150 epochs & 128 & $1\times10^{-4}$ 
& None & 0.9 & 0.001 \\

ResNet18 & Tiny-ImageNet & Finetune
& 150 epochs & 128 & $1\times10^{-3} \rightarrow 1\times10^{-5}$ 
& Cosine & 0.9 & 0.001 \\
\midrule

MobileNetV2 & ADE20K & GELU phase 
& 190 epochs & 8 & $0.005 \rightarrow 1\times10^{-4}$ 
& PolyLR & 0.9 & $5\times10^{-4}$ \\

MobileNetV2 & ADE20K & Switch to ReLU 
& 100 epochs & 8 & $1\times10^{-4}$ 
& None & 0.9 & $5\times10^{-4}$ \\

MobileNetV2 & ADE20K & Finetune 
& 100 epochs & 8 & $1\times10^{-4} \rightarrow 5\times10^{-6}$ 
& Linear & 0.9 & $5\times10^{-4}$ \\
\bottomrule
\end{tabular}
\end{adjustbox}
\end{table*}

\section{Layer Sharing Details}
\label{sec:layer_sharing}
In the method section, we defined
\[
    \mathrm{sharedReLU}(x^{c}) = 
    x^{c} \cdot 
    \left(\alpha^{c} \cdot \mathrm{DReLU}(x^{\pi(c)}) + \beta^{c}\right),
\]
where $\pi$ is a mapping from each replicate channel to its corresponding prototype channel (and from a prototype channel to itself).

We now extend this formulation to the setting of \emph{layer sharing}.  
Let $\pi^{l}$ denote the channel–mapping function for layer $l$.  
To enable sharing across layers, we partition the network's layers into groups of consecutive layers that share the same spatial and channel dimensions.  
We then define a function
\[
    \phi : \{1,\ldots,L\} \rightarrow \{1,\ldots,L\},
\]
where $L$ is the total number of layers, and $\phi(l)$ returns the index of the first layer in the group containing layer $l$.

We denote by $x^{l,c}_{h,w}$ the neuron value at layer $l$, channel $c$, and spatial location $(h,w)$.  
For notational convenience, we set
\[
    \pi^{l} = \pi^{\phi(l)},
\]
meaning that all layers within the same group use the channel–mapping function of the group's first layer.

Under this construction, the layer–shared activation becomes
\begin{equation}
\begin{aligned}
\mathrm{sharedReLU}(x^{l,c}_{h,w})
&= x^{l,c}_{h,w} \cdot 
\Big(
    \alpha^{l,c} \cdot 
    \mathrm{DReLU}\!\left(
        x^{\phi(l),\pi^{l}(c)}_{h,w}
    \right) \\[-0.2em]
&\qquad\qquad
    + \beta^{l,c}
\Big).
\end{aligned}
\end{equation}

\section{Additional Experimental Figures}
\label{sec:additional_figs}

\begin{figure}[tbp]
    \centering
    \includegraphics[width=0.8\linewidth]{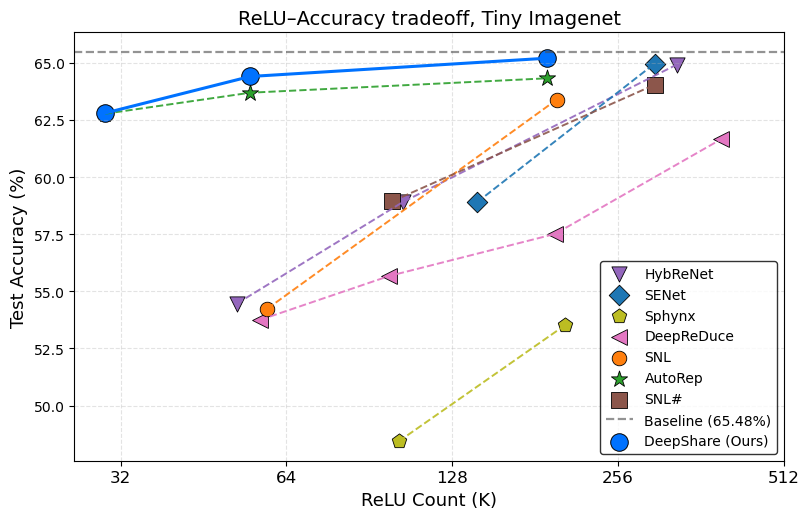}
    \caption{DeepShare achieves the Pareto frontier of ReLU count versus test accuracy on Tiny-ImageNet using ResNet-18.}
    \label{fig:tiny}
\end{figure}

Figure~\ref{fig:tiny} presents the Tiny-ImageNet results, demonstrating that DeepShare attains the Pareto-optimal trade-off between ReLU budget and accuracy.

\section{Analysis of the Affine Parameters}
\label{sec:affine_analysis}
In our method, the gate used by each channel—whether prototype or replicate—is first passed through an affine transformation before being applied to the activation. 
As detailed in the method section, for each channel \(c\), the learned parameters \(\alpha^{c}, \beta^{c} \in \mathbb{R}\) are used in
\[
    \mathrm{sharedReLU}(x^{c}) =
    x^{c} \cdot 
    \left(\alpha^{c} \cdot \mathrm{DReLU}(x^{\pi(c)}) + \beta^{c}\right),
\]
\[
    \mathrm{sharedReLU}(x^{c}) =
    x^{c} \cdot 
    \left(\alpha^{c} \cdot \mathrm{relu\_gate}(x^{\pi(c)}) + \beta^{c}\right),
\]

With $\alpha^c$ acting as the weight and $\beta^c$ as the bias.

We provide an initial examination of how this affine transformation behaves in trained networks.  
Interpreting neural networks from their parameter values is a substantial task in itself and often ambiguous; therefore, our goal here is not to provide definitive conclusions, but rather to highlight preliminary trends and insights observable from the learned parameters.

\begin{figure*}[t]
    \centering
    \includegraphics[width=0.8\linewidth]{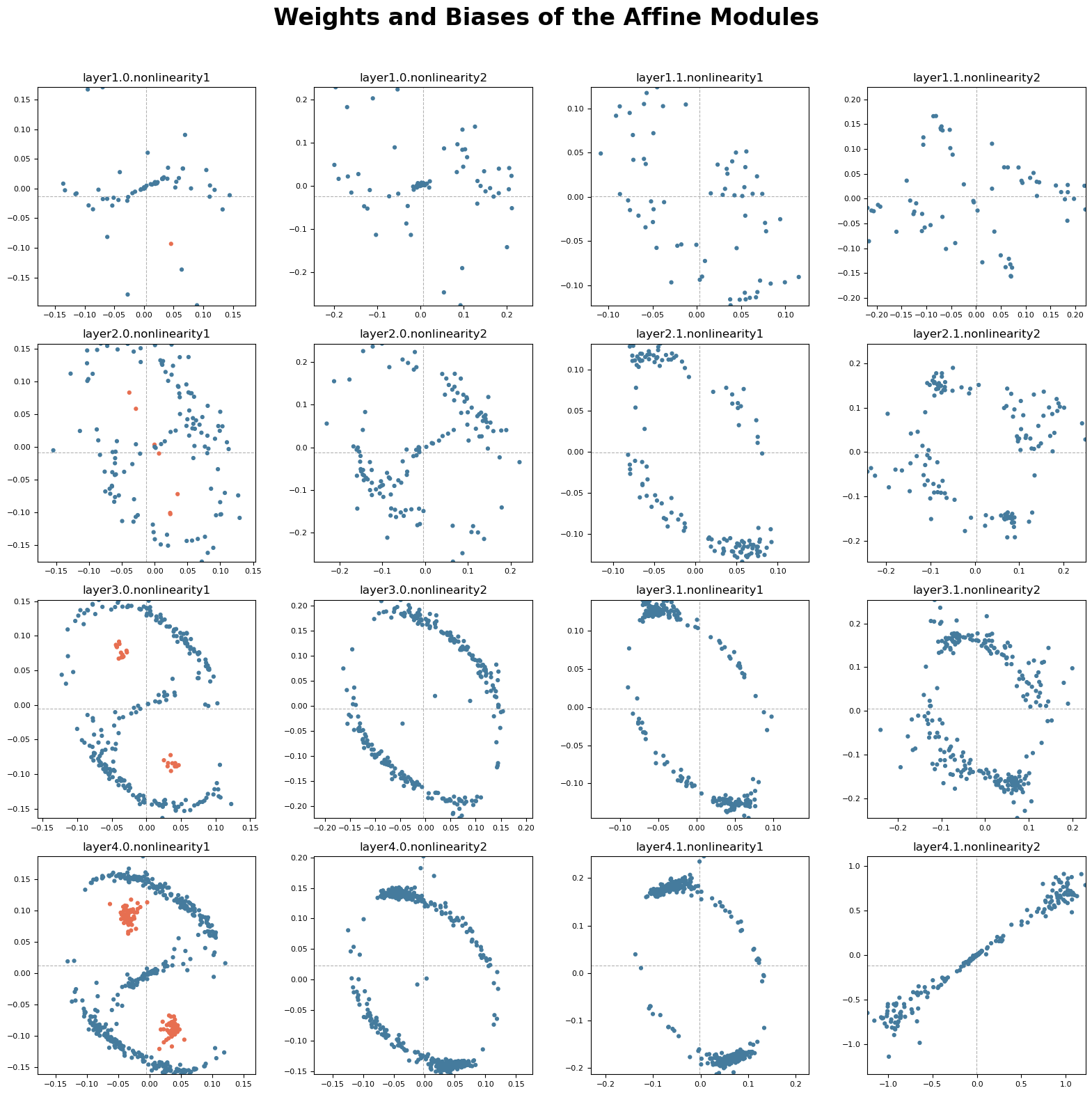}
    \caption{    Affine 
    gate parameters $(\alpha^c,\beta^c)$ for a trained ResNet-18 with a 6K-gate budget. 
Each subplot corresponds to one layer, and each point represents a channel in that layer. 
The horizontal axis shows the bias term $\beta^c$, and the vertical axis shows the weight term $\alpha^c$. 
Prototype channels (orange) appear only in the first layer of each layer-sharing group; replicate channels are shown in blue.
    }
    \label{fig:open_close}
\end{figure*}

\paragraph{Initial Observations}
Figure~\ref{fig:open_close} plots the $(\alpha^c,\beta^c)$ pairs for each channel and layer of a trained model. We observe that similar patterns emerge across different ReLU budgets. \\
Our main observations are as follows:

\begin{itemize}
    \item \textbf{Replicate diversity.}  
    In most layers, the replicate channels disperse across a wide range of angles along an approximately elliptical (or circular) manifold, in clear contrast to their uniform initialization. This spread indicates that the prototype gates are utilized in diverse ways across channels, rather than collapsing to a single dominant pattern or remaining uniformly spread.

    \item \textbf{Prototype structure.}  
    In contrast to the dispersed behavior of the replicate channels, the prototype channels display a clustered behavior. In the deeper layers, they form two symmetric clusters (up to sign) characterized by relatively large weights compared to their biases, suggesting a prominent use of the gate.

    \item \textbf{Linear-like neurons.}  
    Although less common, some channels exhibit very small weight magnitudes, indicating that their outputs are only weakly modulated by the gate. These channels behave similarly to linear units without an activation.
\end{itemize}

\end{document}